\documentclass[conference]{IEEEtran}
\IEEEoverridecommandlockouts
\usepackage{cite}
\usepackage{amsmath,amssymb,amsfonts}
\usepackage{algorithmic}
\usepackage{graphicx}
\usepackage{textcomp}
\usepackage{xcolor}
\begin{document}

\title{Proactive Security: Embedded AI Solution for Violent and Abusive Speech Recognition}

\author{\IEEEauthorblockN{1\textsuperscript{st} Christopher Dane Shulby}
\IEEEauthorblockA{\textit{Bixby Solutions} \\
\textit{Samsung Research SIDI Institute}\\
Campinas, Brazil \\
c.shulby@sidi.org.br}
\and
\IEEEauthorblockN{2\textsuperscript{nd} Leonardo Pombal}
\IEEEauthorblockA{\textit{Security} \\
\textit{Samsung Research SIDI Institute}\\
Campinas, Brazil \\
l.pombal@sidi.org.br}
\and
\IEEEauthorblockN{3\textsuperscript{rd} Vitor Jord\~ao}
\IEEEauthorblockA{\textit{Bixby Solutions} \\
\textit{Samsung Research SIDI Institute, Brazil}\\
Campinas, Brazil \\
v.jordao@sidi.org.br}
\and
\IEEEauthorblockN{4\textsuperscript{th} Guilherme Ziolle}
\IEEEauthorblockA{\textit{Mobile Product} \\
\textit{Samsung Research SIDI Institute, Brazil}\\
Campinas, Brazil \\
g.ziolle@sidi.org.br}
\and
\IEEEauthorblockN{5\textsuperscript{th} Bruno Martho}
\IEEEauthorblockA{\textit{Bixby Solutions} \\
\textit{Samsung Research SIDI Institute, Brazil}\\
Campinas, Brazil \\
b.juliani@sidi.org.br}
\and
\IEEEauthorblockN{6\textsuperscript{th} Ant\^onio Postal}
\IEEEauthorblockA{\textit{RD Support} \\
\textit{Samsung Research SIDI Institute, Brazil}\\
Campinas, Brazil \\
a.postal@sidi.org.br}
\and
\IEEEauthorblockN{6\textsuperscript{th} Thiago Prochnow}
\IEEEauthorblockA{\textit{Mobile Product} \\
\textit{Samsung Research SIDI Institute, Brazil}\\
Campinas, Brazil \\
t.prochnow@sidi.org.br}
}

\maketitle

\begin{abstract}
Violence is an epidemic in Brazil and a problem on the rise world-wide. Mobile devices provide communication technologies which can be used to monitor and alert about violent situations. However, current solutions, like panic buttons or safe words, might increase the loss of life in violent situations. We propose an embedded artificial intelligence solution, using natural language and speech processing technology, to silently alert someone who can help in this situation. The corpus used contains 400 positive phrases and 800 negative phrases, totaling 1,200 sentences which are classified using two well-known extraction methods for natural language processing tasks: bag-of-words and word embeddings and classified with a support vector machine. We describe the proof-of-concept product in development with promising results, indicating a path towards a commercial product. More importantly we show that model improvements via word embeddings and data augmentation techniques provide an intrinsically robust model. The final embedded solution also has a small footprint of less than 10 MB.
\end{abstract}


  \section{Introduction and Motivation}
  
    Technology is under constant evolution and people desire better resources to improve public and personal safety, especially in Latin American countries like Brazil. In Brazil, criminal situations like assaults, vehicle theft and robbery, domestic violence like rape and domestic assaults occur every day. Normally, victims have no time to react and if they do, abrupt movements or odd behavior can provoke the aggressor to intensify his/her level of aggression, resulting in serious injuries or even fatalities. Year after year, the number of violent cases continues to grow, especially in Brazil and similar countries. The number of property thefts in 2012 was 841.663 according to the institute IPEA \cite{CrimesViolentos2017}. This statistic represents a number of 422 incidents that year for every 100 thousand people in Brazil. According to the WHO, Brazil had the highest absolute number of murders in the world in 2012 \cite{globoNews2014} with 47 thousand murders; that represents 13 in each 100 murders which occurred in the entire world.  
    
	Technology can be our ally to improve security. Brazil is a country with large number of active smartphones. According to Anatel \footnote{Brazilian Government Agency of Telecommunications}, in 2017, 243.4 million cellphones were active \cite{Telco2017}. That means approximately 1.18 cellphones/person in the Brazilian population, indicating there are more cellphones than people. Cellphone theft increases the necessity for Brazilians to insure \cite{globoNews2016} their smartphones and pay for backup communication plans to reduce losses in the case of incidents. However it's not a definitive solution. In this scenario, the existing solutions of security alarms are not adequate because they require user interaction and increase the risk of life as a result of the aggressor's reactions. The Proactive Security project was developed as a PoC for BP to solve this issue, sending silent alarms based on human conversations and analyzing, by NLP technology, whether it is a violent case or not. In the case of a violent event, the software will notify a destination defined by the user, sending information about the event so the designated person can decide to call for help.
The main goal is reduce loss of life during violent confrontations.
  
\section{Related Work}

\subsection{Speech Recognition for Violent Incidents}
    There have been some patents with various degrees of relation to the current work, but we were not able to identify a patent for embedded speech recognition for violent utterance identification. We are under continuous analysis of the prior-art and other works about these related solutions. For example, the closest patent we found was US patent application US20160071399A1, a ``Personal Security System" \cite{altman2016personal} that continually monitors audio from any microphone accessible from a computer or mobile phone, waiting for a specific key; once it identifies the key emitted by a user, it issues an alert. Audio monitoring may search for keys in a personalized manner and be utilized by persons unable to physically manipulate their phone or computer, during or after an assault, or medical emergency. It can recognize a keyword of at least one word via application and optionally accept additional intonations for words, phrases or sounds. This solution is similar to ours in terms of application but not strategy. The disadvantage of this approach is that the user must use a special keyword, or safe word.  In cases where the victim is known to the perpetrator, it is likely that the aggressor also knows intimate information about the victim, including such a keyword.  Also, even in cases where this is not true, it is likely that sudden shouting of any specific word could be interpreted as strange behavior, increasing the likelihood that the aggressor would react, possibly fatally. 
    
    In the event of a robbery, the aggressor is typically in a nervous or an altered state and their reactions are not necessarily rational or predictable. When the victim reacts to an armed aggressor, the probability that the victim will be killed is very large because any brusque movement can scare the aggressor. With our solution, no specific keyword is used as a trigger, but instead an intelligent NLP-based solution is proposed.
    
    The next closest related patent is US granted patent 9,805,576, a ``Smart Watch with Automatic Voice Recording and Alarm" \cite{nguyen2017smart}.  In this case a wearable is utilized and a panic signal is triggered by a heart rate monitor, temperatures and also ``filtered voice" from the environment. Voice content is not recognized, just filtered and recorded. This patent is somewhat less related. Heart rate seems a less reliable indicator of physical risk and could simply be due to exercise or outdoor activities. In any case, the method is certainly different from ours since no NLP or ASR solution is present.
    
\subsection{NLP Techniques for Identifying Semantics in Phrases}

\subsubsection{Bag of Words}
    The bag-of-words model is a representation of a set of text (such as sentences or documents), based on its words \cite{sivic2009efficient}. Given a set of texts, such as sentences or documents, the goal of this model is to transform each text into a numeric vector, based on the frequency of each word. Each element of the vector can be used as a feature for a machine learning algorithm, such as a SVM or neural network. Numeric vectors are built using different approaches, as explained in \cite{matsubara2003pretext}:

\begin{itemize}
   \item Boolean: the presence of the word in the feature set. A set of the most frequently used words is built, and then each text is transformed into a vector, based on the word set.
   
   \item tf: the term's frequency in the document. This approach keeps the word's multiplicity. It is presented by:
  
\[ tf(tj, di) = freq(tj, di) \]
  
   \item tfidf: the term's frequency multiplied by a correction factor, based on the number of occurrences of the word in the whole set.
  The correction factor is calculated by the number of documents in the set and the number of documents in which the word can be found:
  
\[
\begin{split}
& tf idf(tj, di) = freq(ti, dj) . idf(tj)\\
&idf(tj) = log (N/d(tj))\\
\end{split}
\]

  Where N is the number of the documents, ti is the i-th term of the sentence and dj is the j-th document in the set
\end{itemize}
    
\subsubsection{Word Embeddings}

	NLP applications usually receive words as basic input units. Therefore, Their meaningful representation is important. In recent years, more sophisticated NLP techniques have been developed. \cite{bengio2003neural} was one of the pioneers in employing neural networks to automatically learn vector representations and more effective techniques have appeared within the last half decade or so \cite{collobertetal2011,mikolovetal2013, Ling:2015:naacl, lai2015recurrent}. This type of representation is known as word embedding. Word embeddings are able to capture semantic, syntactic and morphological information from large unannotated corpora. They are vectors of real valued numbers, which represent words in an $n$-dimensional space. 
    
    In order to apply classifiers in NLP tasks, it is necessary to map words of a text to a numeric vector, or word embedding. This representation allows the classifier to capture latent lexical and semantic relations \cite {collobertetal2011, mikolovetal2013}. Word embeddings are being applied to many syntactic and semantic tasks such speech recognition \cite{mikolov2009neural}, semantic similarity \cite{mikolovetal2013}, PoS tagging, sentiment analysis \cite{li2015multi} and logical semantics \cite{bowman2014recursive}. 
    
	Word2Vec is a widely used method in NLP for generating word embeddings by inducing a dense representation of a word \cite{collobertetal2011}, where the network does not have a hidden layer, resulting in a fast log-linear model \cite{mikolovetal2013}. This network can be divided in two architectures: 1.) CBOW, where given a sequence of words, the model attempts to predict the word in the middle; and 2.) Skip-Gram, where given a word, the model attempts to predict its neighboring words. 

	It has been shown by the original work \cite{mikolovetal2013} that Word2Vec is able to capture semantic information in the induced embeddings. Wang2Vec is a modification of Word2Vec made in order to take into account the lack of word order in the original architecture. Two simple modifications were proposed in Wang2Vec expecting embeddings to better capture syntactic behavior of words \cite{Ling:2015:naacl}. In the CBOW architecture, the input is the concatenation of the context word embeddings in the order they occur. In Structured Skip-Gram, a different set of parameters is used to predict each context word, depending on its position relative to the target word. Then, the network has positional information of a word and its neighbors to induce word embeddings \cite{Ling:2015:naacl}.
    
    In the case of BP the induction method with the most stable results for both semantic and syntactic representations is Wang2Vec with CBOW, as shown by \cite{hartmann2017portuguese}. This study is the most exhaustive analysis of word embeddings for NLP tasks in Portuguese to date and evaluated GloVe, Word2Vec, FastText and Wang2Vec methods, all with both CBOW and Skip-gram methods intrinsically on syntactic and semantic analogies and extrinsically on the PoS Tagging task and on the Semantic similarity task.
    
\subsubsection{Classification with SVM}

	SVM has been shown to be a robust classifier and provides a strong learning guarantee according to the Statistical Learning Theory and large-margin bounds~\cite{vapnik1998statistical, von2008statistical}. The VC theory \cite{vapnik1998statistical} gives us confidence that we are able to generalize new hypotheses well. Also, word embeddings produce large dimensions and the data set is unbalanced. SVM is able to handle large dimensions without creating local distortions like neural networks and statistical data augmentation techniques can be used \cite{shulby2017acoustic}. Since the current paper aims to propose an embedded solution, small models with better generalization seem to be a logical avenue.

\section{Proposed Solution}

\subsection{Dataset}
     
	The dataset was built using real violent and abusive situations, found on websites\cite{gales2017} (public police occurrences, FBI, threat emails, investigation data) and scientific articles \cite{nini2017register}, based on personal experiences. The non-English resources were localized by our localization team to BP. Since the data for BP was rather sparse, some phrases were then augmented using the creative license of the developer team or automatic phrase creation with JSGF scripts. In other words, our team did a brainstorming excercise to best cover violent situations and based on a slotting technique, the database was enlarged. Also, it was necessary to create another dataset, with non-violent phrases, but using some violent or swear words. This dataset was used in training as negative data. It is important that any phrase be analyzed so that the data is not  just for certain words, but rather that the utterance as a whole is evaluated. In total, the dataset consists of 400 positive phrases and 800 negative ones, totaling 1,200 sentences.
    
	With these datasets, an Android application was created to train the model. The datasets were divided into two parts: the training set, composed of a random sample of 70\% of the sentences, used to train the SVM model; and the test set, composed of the remaining 30\%, to validate it. Once the model is trained, it can be stored in a file, since it is an object. Then, this file can be exported to the client application.
    
    The data preparation is done in the follow way: First, all the text is passed into an automatic stemmer, a process of linguistic normalization, in which the radicals are separated and variant forms of a word are reduced to a common form. Then, the stemmed words of each phrase are tokenized and stop words are removed. This helps to combat the sparseness of the data. With these two processes, it is possible to make a vocabulary of the more common words and use a binary matrix. This matrix has the number of positive and negative phrases from the datasets as rows and the number of words in vocabulary as columns. Each row/column is filled with a "0" if a word from the lexicon is not in the phrase or a "1" if is.

\subsection{Architecture and Experiments Proposed}

\subsubsection{General Architecture}

    The solution consists of an Android application embedded or installed in a smart phone. This application will listen to and capture the environment audio and, using an embedded classifier model based on the training set, will classify this audio either as a risk or a non-risk situation. The final embedded solution has a small footprint of less than 10 MB.
    
    The first step is to capture the audio using the smart phone microphone and convert it into text. This text is then tokenized and stemmed, as described below, and the result of this process is the input of the classifier model.
    
    The result of the classification will determine the next actions of the application. In the case of a possibly violent situation, the application will act proactively to ensure assistance to the user. This application should run in the background and act without any action by the user. Also, it must act quietly to avoid calling attention to a person in risk.
    
\subsubsection*{Capture}

	To be able to predict a potentially violent situation, the application needs to capture the audio from the environment and process it using the trained model. Once the application is running uninterruptedly, the process of listening with the microphone should also run uninterruptedly.
    
    The native Android speech recognition service, SpeechRecognizer, was used to collect the audio and transcribe it into text. SpeechRecognizer is an Android service designed to listen to the user for a while. The service must be started and then finished after listening to the speech. If the service is started but not finished, the Android OS will eventually kill this process and it must be started again by the application. Since the goal is to listen the to user uninterruptedly, this behavior was modified to avoid that the service is killed by the OS. After collecting and transcribing the audio into text, this text is processed and sent to the model to be classified.

\subsubsection*{Analysis}

	The text is processed the follow way: First, the text input is tokenized and stemmed with the same techniques used in training set. It is then classified using one of the techniques (Bag of Words with SVM or Word Embeddings with SVM), described respectively in details in sections \ref{bow} and \ref{embed}, where the input phrase is classified in a violent or non-violent situation.

\subsubsection*{Communication of Events}

	Once a potentially violent event is detected, the application should act proactively to ensure assistance to the user as fast as possible. In order to do so, the application allows users to determine some ways to inform a third person when they are in danger.
    
    When the application is first installed, the user can set an email account and a phone number to be informed in the case of possible danger. The application will then, in the background, send an SMS and/or an email with a predetermined message to those contacts as soon as it identifies the risk. It will also send the users location (taken from Google Maps), two pictures taken by the front camera and rear camera and a 10-second video recorded after the risk was identified.
    
\subsubsection{Method 1: Bag of Words + SVM}  
\label{bow}
    This model uses the bag-of-words method and SVM to create a classifier for violent situations. The boolean technique was used to create the feature vectors.
    
	Firstly, the dataset was split into two parts: the training set, used to train our SVM classifier, and the test set, to validate it.  Within the training set, the frequency of each word was calculated. The n-most used words were selected to create a feature set. The set's size was chosen to reduce the number of features while maintaining its quality. Then, each sentence in the training set was converted into a numeric vector representing the features of the SVM classifier.

	The SVM classifier was implemented in JAVA, using LIBSVM, which is a well-known and used library. It has two classes (risk, non-risk), and used a RBF kernel, which produced the best results. Its cost parameter was set to 10, and its normalization parameter was set to 0.1. These parameters were found empirically, using a python script which calculated the best-fitting parameters based on the training set. Once the classifier's parameters and kernel were selected, the training set, which was already labeled, was used to train the classifier.

	Once the training phase was completed, the test set was used to calculate the number of true positives, true negatives, false positives and false negatives. With those values, the classifier's accuracy and F1 score were calculated to determine its performance.
    
    \subsubsection{Method 2: Word Embeddings + SVM + SMOTE}
    \label{embed}
    
    The proposed method takes advantage of the machine learning techniques used for Word Embeddings and SVM and attempts to improve accuracy with a minimalistic approach to data augmentation with SMOTE. The features produced by the embeddings were classified using a SVM. 
    
    We Used the Wang2Vec induction method because we believe that both semantic and syntactic features are useful for classifying violent phrases. After empirical tests, the final architecture consisted of the CBOW induction type with 300 dimensions. We could have used a higher dimension size, but since our goal is to embed the model, it made sense to use a more minimalistic approach where data was still being well represented.
    
    We made some small modifications in the SVM parameters for the embeddings model which were found empirically after several experiments. The selected kernel for final experiments was a $4$\textsuperscript{th} order polynomial kernel with $coef0=1$ (as a non-homogeneous kernel) and a cost $C=50$. For this task, namely semantic clues from lexical features, we have a problem with the unbalanced nature of the dataset, a common characteristic for most speech corpora. Since we have more non-violent phrases, than violent ones, it was necessary to use SMOTE~\cite{chawla2002smote}, a data augmentation technique. The synthetic creation of minority samples allows us to treat the classification task with more confidence.

\section{Results}

\begin{figure}[!ht]
\centering
\includegraphics[width=0.4\textwidth]{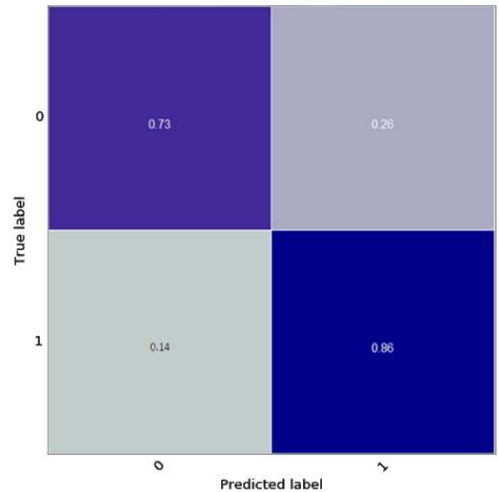}
\caption{{\it Confusion Matrix for Method 1: Bag of Words + SVM}}
\label{fig:BOW}
\end{figure}

	First, we will present the results for the bag-of-words method. The total accuracy on the validation set was 79\% and it achieved an F1 score of 0.78. In Figure~\ref{fig:BOW}, the confusion matrix for the bag-of-words method is presented, where the true-positive rate was 73\%; the false-positive rate was: 26\%; the true negative rate was: 86\%; and the false-negative rate was: 14\%.

\begin{figure}[!ht]
\centering
\includegraphics[width=0.4\textwidth]{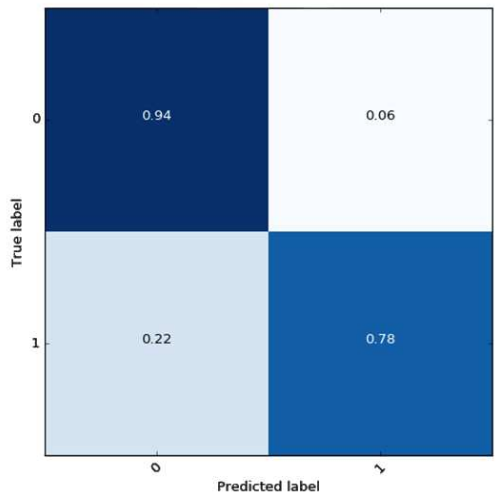}
\caption{{\it Confusion Matrix for Method 2: Embeddings + SVM + SMOTE}}
\label{fig:embeddings}
\end{figure}

Finally, we will present the results for the embeddings method. The total accuracy on the validation set was 87.5\% and it achieved an F1 score of 0.87. In Figure~\ref{fig:embeddings}, the confusion matrix for the word embeddings method is presented, where the true-positive rate was 94\%; the false-positive rate was: 6\%; the true negative rate was: 78\%; and the false-negative rate was: 22\%.

\section{Discussion and Conclusion}
    
    While the first method did produce good results, we were mainly concerned with the high number of false positives at 26\%. The word embeddings were able to better represent the data and SMOTE also helped fix the unbalanced data problem.  In this experiment, the false positive ratio was reduced to 6\% as can be seen in Figure~\ref{fig:embeddings}. The true negative mark was slightly worse in the embeddings model compared with the BOW model.  This is probably from the influence of data augmentation where the first model may have been influenced by the larger dataset of negative phrases. These results are very good because it shows that we are more likely to identify a violent phrase correctly than anything else at 94\%, which is a classic anecdote among machine learners.
    It should also be noted that the F1 score and the accuracy are very close, showing that the model is able to generalize unseen examples almost just as well as the trained examples, demonstrating the intrinsic robustness of this model. We were unable to track extrinsic evaluations, but anecdotally, the solution has been presented in multiple internal and external demos and received much praise and positive reviews. This is important as the solution is currently in a pre-commercialization phase and tests by B2B clients.
    
\section{Future Work and Language Expansion}

	For future work, a deeper system integration is required, in order to enhance the way the environment's audio is captured. The use of the application level API, the SpeechRecognizer Android feature was required for this proof-of-concept but, since this API was not designed to run uninterruptedly, another technique should be used. First of all, it requires muting the phone media because, by default, this feature makes a "beep" sound, always when it starts listening, which is contradictory to the requirement of a silent application. Moreover, in order to avoid that the Android OS kills the service and consequently stops the process, it restarts the service every time it is killed, causing a lot of battery consumption. Currently, we were able to simulate lives tests showing a battery duration of about 15 hours under continuous use. This could possibly be resolved by Samsung's native wake-up or some other intelligent strategy.

	Another future task is to expand the training set. We have since collected a much larger database with different violent and abusive phrases, many of them directly from local police who have indicated an interest in the service. This expansion should improve the accuracy of the model. Also, it is possible to test other machine learning techniques and test the application in the real world to see if it's useful to the user. It is important that we conduct extrinsic, task-based tests, to show the true marketability of the product. As far as features, the second phase of the project plans to include prosodic features in the classification since the tone of voice and speech rhythm are normally distinct in violent and non-violent situations. This would further avoid false positives.
    
    We also plan to integrate this solution in other electronic devices, using IoT technologies. It would allow the solution to be more robust and precise, and use other resources than those provided by the smart phone, such as security cameras, alarm systems, etc. An integration with appliances would be extremely helpful to detect cases of domestic violence, for instance.
    
    The solution should also be expanded to multiple languages, since until now it's been developed only for BP, creating new training sets for each scenario and applying the same model for identifying violent and abusive speech. This means that the algorithm is language independent and could be used for other languages with new data. If necessary, the word embeddings methods could be revised for some non-European languages, but this may not be necessary. 
    
\section{List of Abbreviations}
The following abbreviations and acronyms were used in this paper:\\~\\
API (Application Programming Interface)\\
BP (Brazilian Portuguese)\\
tf (Term Frequency)\\
tfidf (Term Frequency Inverse Document Frequency)\\
IPEA (Instituto de Pesquisa Econômica Aplicada)\\
IoT (Internet of Things)\\
JSGF (Java Speech Grammar Format)\\
NLP (Natural Language Processing)\\
OS (Operating System)\\
RBF (Radial Basis Function)\\
PoC (Proof of Concept)\\
PoS (Part of Speech)\\
SVM (Support Vector Machine)\\
VC (Vapnik–Chervonenkis)\\
WHO (World Health Organization)\\


\section{Acknowledgements}
	We would like to thank our SIDI colleagues for their input and support...

\bibliographystyle{IEEEtran}

\bibliography{mybib}

\end{document}